\documentclass[11pt]{article}
\usepackage[preprint]{acl}
\usepackage{times}
\usepackage{latexsym}
\usepackage[T1]{fontenc}
\usepackage[utf8]{inputenc}
\usepackage{microtype}
\usepackage{inconsolata}
\usepackage{graphicx}
\usepackage{booktabs}
\usepackage{amsmath}
\usepackage{amssymb}
\usepackage{multirow}
\usepackage{xcolor}
\usepackage{enumitem}
\setlist{nosep,leftmargin=*}
\graphicspath{{figures/}}
\newcommand{\laya}{\textsc{Laya}}
\newcommand{\jev}{\textsc{Jev}}
\newcommand{\pp}{\,pp}
\newcommand{\dec}[1]{\texttt{#1}}
\providecommand{\Description}[1]{}
\providecommand{\figw}{\columnwidth}
\providecommand{\figwide}{\textwidth}
\newcommand{\codeurl}{\url{https://github.com/David-DL-Space/sys1-eval}}

\title{Fast Models, Slow Evidence:\\ A Paired and Self-Audited Evaluation of System-1 Decision Models for LLM Agent Harnesses}
\author{Jiawei Li}

\begin{document}
\maketitle
\begin{abstract}
Agent harnesses make many small, typed decisions per task: which model to call, which tool to use, whether retrieved text is relevant, whether an input carries an injection. \emph{System-1 decision models} answer such questions in a single forward pass with class probabilities, promising large cost and latency savings over LLM calls. We present a paired evaluation of an open-weight (\laya{}) and a hosted (\jev{}) System-1 model on 11 agent decision points built from 18 public sources: 7,283 base cases plus 6,640 robustness variants, with byte-identical inputs, paired tests, and cross-hardware and cross-day reproducibility checks. \jev{} is significantly more accurate on 9 of 11 decision points (+10.8 to +46.0\pp). Neither model beats chance on zero-shot model routing, and they tie on RAG relevance gating. \laya{} changes 30\% of its answers when the option order is reversed and degrades sharply with many or similar candidates (31\% at 50 nearest-neighbour tools, vs.\ 98\% for \jev{} on items with a unique correct tool). We also audit our own pipeline. Three analysis errors and one design confound distorted headline deployment claims: an omitted pre-screen cost (reported 23.9\% saving, actual 4.3\%), gate accuracy reported as end-to-end quality (58\% vs.\ 98\%), in-sample thresholds (5\% target, up to 17\% held-out misses), and a ``channel effect'' on injection false positives that vanishes with channel-native content. Two other suspected confounds did not change the conclusions. All cases, raw outputs and analysis code are available at \codeurl{}.
\end{abstract}

\section{Introduction}
An LLM agent is surrounded by a \emph{harness}: code that routes requests, selects tools, decides whether to call a tool at all, gates retrieval, and screens untrusted content. Each of these is a \emph{decision}: a typed output from a small, known set. Most of them are currently made by prompting the same LLM that does the work, which spends hundreds of milliseconds and a full generation on a problem whose output is one of a few labels.

A recent class of models targets exactly this layer. Following \citet{kahneman2011thinking}, their vendors call them \emph{System-1} models: given a \emph{state} and a set of typed \emph{questions} (\dec{choice} among $K$ options, a yes/no probability, or an ordinal score), they return a distribution over answers in one non-autoregressive pass. \laya{} \citep{laya2026} ships open weights; \jev{} \citep{jev2026} is a hosted API. Both claim LLM-level accuracy on structured decisions at a small fraction of the cost.

Practitioners who want to adopt such a model need answers to three concrete questions. \textbf{(RQ1)} At which harness decision points is a System-1 model accurate enough? \textbf{(RQ2)} How reliable is it: stable under option order, wording, input channel, and model version, and calibrated enough to threshold? \textbf{(RQ3)} When used as a \emph{gate} in front of an LLM, what does it actually save, and at what risk?

We answer these with a paired evaluation (\S\ref{sec:setup}) and make three contributions:
\begin{itemize}
\item \textbf{A decision-point benchmark for agent harnesses.} It has 11 decision points, from model routing to PII egress, built from 18 public sources, with byte-identical inputs to both systems and paired statistics (\S\ref{sec:bench}).
\item \textbf{Findings for practitioners} (\S\ref{sec:results}, \S\ref{sec:gates}). \jev{} dominates on tool selection, large label sets and injection screening. Both systems fail at zero-shot routing and tie on RAG gating. \laya{} is order-sensitive, collapses with similar candidates, and misses half of the injections embedded in documents. Used as gates, System-1 models save far less than naive accounting suggests.
\item \textbf{A self-audit of the evaluation itself} (\S\ref{sec:audit}). The first version of this study contained errors that changed deployment recommendations. We quantify each error, run control experiments for each suspected confound, and report two that turned out to be harmless. We believe this checklist transfers to any evaluation of gating components.
\end{itemize}

\section{Related Work}
\paragraph{Decision layers in agents.}
Cost-aware routing and cascades send easy queries to cheaper models \citep{chen2023frugalgpt,ong2024routellm}. RouterBench provides per-query outcomes of many LLMs for studying such routers \citep{hu2024routerbench}. Tool use is evaluated by BFCL, including \emph{irrelevance} cases where no tool should be called \citep{patil2023gorilla,patil2025bfcl}. Adaptive retrieval decides when to retrieve or trust context \citep{asai2024selfrag,yan2024crag,jeong2024adaptiverag}. Safety classifiers such as Llama Guard screen inputs and outputs \citep{inan2023llamaguard}. All of these are typically implemented either as a trained task-specific classifier or as an LLM prompt. System-1 models sit in between: one general model, zero-shot over typed questions.

\paragraph{Prompt injection.}
Indirect prompt injection places adversarial instructions in data that the agent retrieves \citep{greshake2023indirect}. AgentDojo, InjecAgent and BIPIA evaluate agents and defenses under such attacks \citep{debenedetti2024agentdojo,zhan2024injecagent,yi2023bipia}. We evaluate the \emph{detector} in isolation and study how the delivery channel interacts with content style.

\paragraph{Evaluation methodology.}
We use paired tests \citep{mcnemar1947note,dror2018hitchhiker}, Wilson intervals \citep{wilson1927probable}, calibration metrics \citep{guo2017calibration,platt1999probabilistic}, and selective-prediction views of thresholding \citep{geifman2017selective}. Order and wording sensitivity of LLM classifiers is well documented \citep{zheng2024mcq,pezeshkpour2024order,sclar2024quantifying}. So are label errors in test sets \citep{northcutt2021pervasive}, contamination of public benchmarks \citep{sainz2023contamination}, and the limits of LLM judges \citep{zheng2023judging}. Our self-audit adds a less discussed failure class: errors in \emph{deployment accounting} (cost, end-to-end quality, threshold selection) that sit downstream of correctly scored predictions.

\section{Systems and Protocol}\label{sec:setup}
\paragraph{Interface.}
Both systems accept a \emph{state} (here always a string) and a dictionary of \emph{questions}. A \dec{choice} question has \dec{instructions} and a \dec{criteria} map from option keys to short descriptions. A yes/no question (\dec{noul} in both APIs) returns $p(\text{yes})$. We use \dec{score} questions only as auxiliary signals. \laya{} 0.3.21 routes each request to an English or a multilingual checkpoint. \jev{} returned the version string \texttt{jev-1.13.0} on every call we made, across two days.

\paragraph{Fairness protocol.}
Both systems receive byte-identical \dec{state} and \dec{questions} for every case. Confidence is $\max_k p_k$ for \dec{choice} and $\max(p,1-p)$ for yes/no. This follows \laya{}'s own calibration code. Its response field \texttt{confidence} is an entropy-based quantity on a different scale: it differs from $\max_k p_k$ by 0.299 on average, versus 0.010 for \jev{}. Using that field would silently break any threshold. Accuracy intervals are Wilson; model comparisons use the exact McNemar test on discordant pairs, with an asymptotic CI for the paired difference. We do not correct for multiplicity. Every claim of a difference below has $p<0.001$, and we call $p>0.4$ a tie only where the CI is narrow.

\paragraph{Determinism and reproducibility.}
Re-running 420 cases, \laya{} changed 0 decisions and \jev{} changed 1. On a second machine (Apple M4 CPU vs.\ an RTX 2080 Ti GPU), \laya{} reproduced 1,180/1,180 decisions (max $|\Delta p|=0.006$). One day later, \jev{} changed 5 of 1,080 decisions (0.5\%). Pairing runs across days and machines is therefore safe for both systems.

\paragraph{Reference LLM.}
For the control experiments (\S\ref{sec:audit}) we add DeepSeek-V4.1-Flash \citep{deepseek2026flash} as a reference. It is prompted with the same state and question (temperature 0; prompt in App.~\ref{app:prompts}). It is \emph{not} part of the paired comparison: its input format necessarily differs.

\section{Benchmark}\label{sec:bench}
Each decision point is a question a harness asks, instantiated from public data (Table~\ref{tab:bench}; details in App.~\ref{app:data}). The 2,100-case suite (A--F) is disjoint from a 100-case pilot whose labels were checked case by case. The agent-infrastructure suite (G--J, 1,280 cases) uses paired constructions wherever possible: the same payload in two channels, the same evidence with a correct and a hallucinated answer, and the same text with and without PII.

\begin{table}[t]
\centering\footnotesize
\setlength{\tabcolsep}{2.5pt}
\resizebox{\figw}{!}{\begin{tabular}{@{}llrl@{}}
\toprule
Decision point & Source(s) & $n$ & Type\\
\midrule
A Model routing & RouterBench & 400 & ch(2)\\
B Tool selection & BFCL & 400 & ch($K$)\\
C Tool-call guard & BFCL + irrelevance & 300 & y/n\\
D RAG relevance & BEIR$\times$5 + BM25 & 1,920 & y/n\\
E Injection guard & deepset, jackhhao & 400 & y/n\\
F Large label set & BANKING77, MASSIVE & 200 & ch(60+)\\
G Injection$\times$2 ch. & deepset, jackhhao & 480 & y/n\\
H Groundedness & HaluEval-QA & 200 & y/n\\
I Pre-exec.\ safety & BeaverT., HarmB., DNA & 400 & y/n\\
J PII egress & ai4privacy & 200 & y/n\\
\midrule
B2 Hard distractors & BFCL & 400 & ch($K$)\\
G2 Inj.\ in documents & 3 corpora + AgentDojo & 480 & y/n\\
J2 Natural redaction & ai4privacy & 100 & y/n\\
S Two-tier safety & I sources + Alpaca & 603 & y/n\\
\bottomrule
\end{tabular}}
\caption{Decision points. B2, G2 and J2 are the control experiments of \S\ref{sec:audit}. S adds a policy-labelled gold standard for \S\ref{sec:gates}. $K\in\{2,5,10,20,50\}$ with 80 cases per value. ch = choice, y/n = yes/no.}
\label{tab:bench}
\end{table}

\paragraph{Label construction.}
For A, a request is labelled \emph{cheap} if both a 7B model and GPT-4 solved it in RouterBench, and \emph{strong} if only GPT-4 did; items that neither solved are dropped. For B, the correct tool is BFCL's gold function, and $K{-}1$ distractors are drawn at random from the pool of all BFCL functions (B2 replaces them with lexical nearest neighbours, \S\ref{sec:audit}). For D, positives are BEIR relevance judgements \citep{thakur2021beir}; negatives are BM25 hard negatives (800) or random passages (320). Other tracks use the dataset labels. We quantify their noise in \S\ref{sec:threats}.

\section{Results: Accuracy and Reliability}\label{sec:results}
\begin{figure}[t]
\centering
\includegraphics[width=\figw]{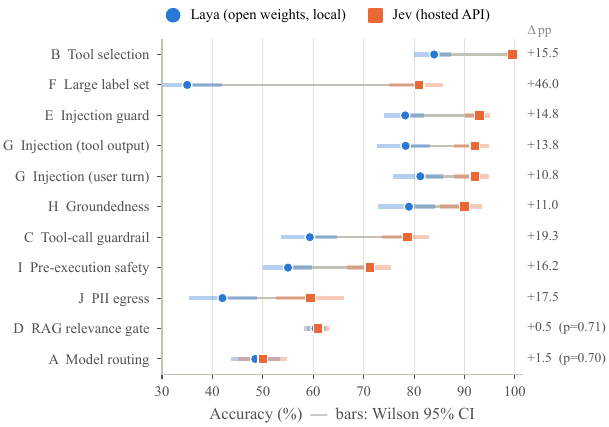}
\Description{Dot plot of accuracy per decision point for Laya and Jev with 95\% intervals; Jev is higher on 9 of 11 points, ties on RAG gating and model routing.}
\caption{Accuracy per decision point. Right: paired difference \jev{}$-$\laya{} (pp); $p<0.001$ unless stated.}
\label{fig:main}
\end{figure}

\paragraph{RQ1: \jev{} is more accurate at 9 of 11 decision points; two are ties.}
Figure~\ref{fig:main} shows paired accuracies. Overall, \jev{} reaches 76.7\% vs.\ 63.1\% on the 2,100-case suite (455 vs.\ 169 discordant wins) and 80.2\% vs.\ 66.0\% on agent infrastructure. The largest gap is on large label sets (F: 81.0\% vs.\ 35.0\%, +46.0\pp{} [38.7, 53.3]).
Two points are ties, for different reasons. \emph{Model routing} (A) is a tie at chance: \jev{} answers ``cheap'' for 99.5\% of requests (50.0\%, the balanced-set constant baseline), and \laya{} has AUROC 0.48. \emph{RAG relevance} (D, $n{=}1{,}920$) is a well-powered tie: 61.0\% vs.\ 60.5\%, difference +0.5\pp{} with CI [$-2.0$, $3.0$]. The two systems err in opposite directions, though. \jev{} recovers more relevant passages (58.3\% vs.\ 45.4\%), and \laya{} rejects more BM25 hard negatives (61.8\% vs.\ 49.3\%).

\paragraph{Error direction matters for guards.}
On the tool-call guardrail (C), \laya{} says ``call a tool'' for 74.7\% of requests that need none (\jev{}: 40.0\%). On pre-execution safety (I), \laya{} lets through 81.5\% of harmful requests (\jev{}: 25.5\%). On PII egress (J), \laya{}'s AUROC is 0.41, i.e.\ its scores rank the classes the wrong way round. We return to J in \S\ref{sec:audit}.

\begin{figure}[t]
\centering
\includegraphics[width=\figw]{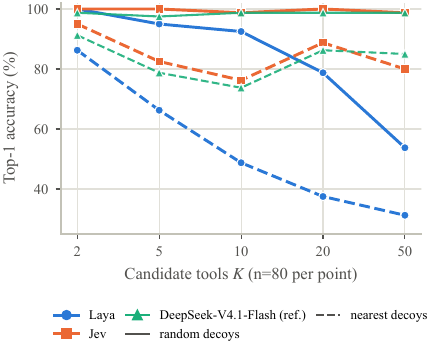}
\Description{Line chart of tool-selection accuracy versus number of candidate tools (2 to 50) for random and nearest-neighbour distractors; Laya drops to 31\% at 50 similar tools while Jev stays above 76\%.}
\caption{Tool selection vs.\ number of candidates. Solid: random distractors (track B). Dashed: nearest-neighbour distractors (B2).}
\label{fig:k}
\end{figure}

\paragraph{Capacity: many and similar candidates.}
With random distractors (Fig.~\ref{fig:k}, solid), \laya{} degrades smoothly with $K$ (100\% at $K{=}2$, 53.8\% at 50), while \jev{} stays at $\geq$98.8\%. The released \laya{} code explains the shape: all options share a 192-token budget, and each option is truncated to $\max(4, (192-16)/K)$ tokens. At $K{=}50$ an option is about four tokens long. With \emph{similar} distractors (dashed), \laya{} falls to 54.0\% overall and 31.3\% at $K{=}50$. \jev{} drops to 84.5\%, but the drop is almost entirely label ambiguity: an LLM audit found another valid tool in 19.8\% of B2 items. On the 70.2\% of items with a unique correct tool, \jev{} scores 98.9\% and \laya{} 61.6\%, a gap of +37.4\pp{} (vs.\ +15.5\pp{} with random distractors). The same pattern holds for multilingual intents (App.~\ref{app:full}): \laya{} scores 0/10 in Arabic and Hindi with 60 labels.

\begin{figure*}[t]
\centering
\includegraphics[width=\figwide]{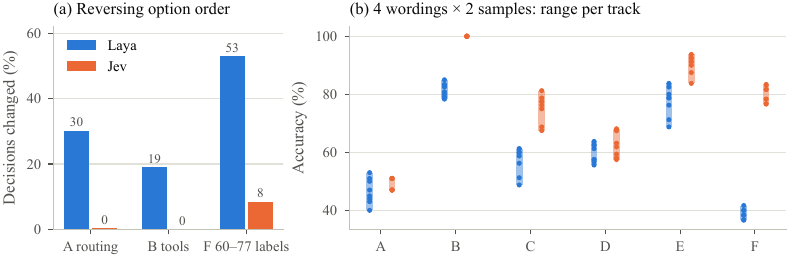}
\Description{Left: bar chart of decisions changed when option order is reversed (Laya 19-53\%, Jev 0-8\%). Right: accuracy ranges across 4 wordings and 2 samples per track.}
\caption{(a) Share of decisions that change when only the option order is reversed ($n=1{,}000$). (b) Accuracy of each track under 4 instruction wordings $\times$ 2 case samples (600 cases each); dots are the 8 runs.}
\label{fig:robust}
\end{figure*}

\paragraph{RQ2: \laya{} is order-sensitive; wording moves both systems.}
Reversing the option list changes 30.3\% of \laya{}'s decisions (53\% with 60--77 labels), compared with 1.8\% for \jev{} (Fig.~\ref{fig:robust}a). In routing, \laya{} picks ``cheap'' for 58.5\% of requests in the original order and 82.3\% in reversed order. Its preference follows position, not content. Rewording only the instructions (criteria fixed) and re-sampling cases moves accuracy by 5--15\pp{} per track for \emph{both} systems (Fig.~\ref{fig:robust}b). Any single accuracy number in this paper therefore carries a wording band of roughly $\pm$5\pp{}. The \emph{ranking} is robust: \jev{} leads in 8/8 wording$\times$sample combinations on B, C, E and F (smallest margin +1.3\pp{} on E). On A and D the sign flips (6/8 and 7/8).

\paragraph{Calibration.}
Pooled ECE is 0.158 (\laya{}) vs.\ 0.143 (\jev{}) on A--F and 0.198 vs.\ 0.080 on G--J (App.~Fig.~\ref{fig:rel}). \jev{}'s A--F figure is dominated by routing, where it is confidently constant (ECE 0.477). Among \laya{}'s answers with confidence $\geq$0.9, 26\% are wrong. \laya{} checkpoints also warn at load time that some temperatures are invalid and ``confidence \ldots{} uncalibrated''. Per-domain recalibration helps: on D, Platt scaling fitted on half the data reduces \laya{}'s ECE on the other half from 0.22 to 0.027 (\jev{}: 0.21 to 0.10).

\paragraph{Latency and cost are deployment properties.}
\laya{} runs in 31\,ms (p50) on an RTX 2080 Ti but 200\,ms on an Apple M4 CPU. \jev{} took 325\,ms (p50) from one client site and 83\,ms from another. The ``10$\times$ faster'' comparison that the first version of this study reported was a property of one network path. \jev{} uses 759 tokens per decision on average, i.e.\ $\approx$\$32 per million decisions at \$0.042/M tokens. \laya{}'s marginal cost is local compute.

\section{Using System-1 Models as Gates}\label{sec:gates}
A \emph{gate} decides whether an expensive path (an LLM call, a review, a block) runs. Two quantities matter: the share of traffic that still takes the expensive path (\emph{call rate}), and the share of cases that needed the expensive path but skipped it (\emph{miss rate}, FNR).

\begin{figure*}[t]
\centering
\includegraphics[width=\figwide]{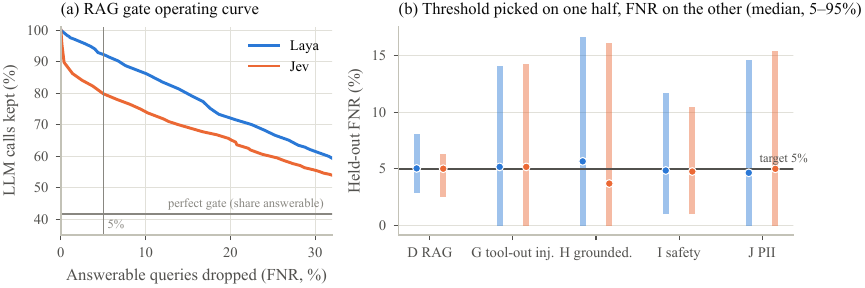}
\Description{Left: RAG gate operating curves of LLM calls kept versus answerable queries dropped. Right: held-out false-negative rates of 5\%-target thresholds with 5-95\% ranges up to 17\%.}
\caption{(a) RAG gate: share of LLM calls kept vs.\ answerable queries dropped (in-sample). (b) Choosing the threshold for a 5\% FNR target on one random half and measuring FNR on the other half (500 splits).}
\label{fig:gates}
\end{figure*}

\paragraph{RAG gating saves little because of base rates.}
If the gate skips generation when context is judged irrelevant, then at a 5\% miss budget \laya{} keeps 92.4\% of LLM calls and \jev{} 79.8\% (Fig.~\ref{fig:gates}a). End-to-end, 97.9\% of answerable queries are still answered by both (assuming the LLM is right). The first version of this study reported 45.1\%/57.7\% here, which is the gate's own classification accuracy (see \S\ref{sec:audit}). The binding constraint is instead the ceiling on savings, which is set by the share of irrelevant retrievals in traffic. In our set a perfect gate saves at most 58\%. Our mix is synthetic, so absolute savings will differ in production. The ranking will not: \jev{}'s AUROC is 0.71 vs.\ 0.61. On real top-5 contexts (400 cases), both systems recognise under 40\% of ``no answer here'' contexts, which makes a skip-gate unsafe. We recommend using the gate as a \emph{trigger for more retrieval}, not as a skip.

\paragraph{Thresholds must be chosen out of sample.}
Picking the threshold that meets a 5\% FNR on the same data it is evaluated on guarantees 5\% by construction. When the threshold is picked on one half and evaluated on the other (Fig.~\ref{fig:gates}b), the median FNR stays near 5\%, but the 95th percentile is 6--8\% for D ($n_{\text{pos}}{=}800$) and 10--17\% for the 100--200-positive tracks. The target is exceeded in roughly half of all splits. A 5\% promise needs either several hundred positives for calibration or a conservative margin.

\begin{figure}[t]
\centering
\includegraphics[width=\figw]{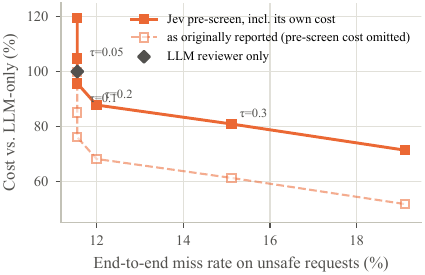}
\Description{Cost relative to LLM-only versus end-to-end miss rate for a Jev pre-screen before an LLM reviewer; corrected costs are 4-20\% lower than LLM-only, not 24-48\% as originally reported.}
\caption{Two-tier safety review (\jev{} pre-screen $\rightarrow$ LLM reviewer): cost vs.\ end-to-end miss rate over thresholds $\tau$. Dashed: the original accounting, which omitted the pre-screen's own cost.}
\label{fig:twotier}
\end{figure}

\paragraph{Two-tier safety review.}
Track S asks whether a System-1 pre-screen can spare an LLM reviewer (DeepSeek-V4.1-Flash) from seeing clearly benign requests. Gold labels come from a written policy applied by a different LLM (GLM-4.6; App.~\ref{app:safety}). This gold agrees with the dataset labels on only 72.8\% of 540 items. At $\tau{=}0.1$, the \jev{} pre-screen sends 65.6\% of traffic to the reviewer with an identical in-sample miss rate (11.6\%). Once the pre-screen's own tokens are counted, however, the cost falls by only 4.3\% (Fig.~\ref{fig:twotier}). At $\tau{=}0.05$ the two-tier system costs 4.7\% \emph{more} than the reviewer alone. The LLM reviewer here is cheap (\$72 per million requests, batched); a single System-1 call is 20\% of that. The durable benefit is \emph{latency}: 34--41\% of requests are released in about 0.3\,s instead of waiting for a multi-second LLM review. With thresholds chosen on held-out halves, the pre-screen adds at least one extra miss in 47\% of splits. \laya{} is useless as a pre-screen: it must pass 99.3\% of traffic to preserve recall.

\paragraph{Cascades between the two System-1 models.}
Letting \laya{} answer when confident ($\geq$0.9) and escalating the rest to \jev{} keeps 41.7\% of decisions local and saves 45\% of \jev{} spend, but it lowers accuracy from 76.7\% to 70.7\%. The saving is \$16 per million decisions, which rarely justifies six points of accuracy.

\section{Auditing the Evaluation}\label{sec:audit}
The first version of this study was built quickly, and its reports were confident and detailed. We re-derived every number from raw outputs and designed a control experiment for every suspected confound. Table~\ref{tab:audit} lists what changed.

\begin{table*}[t]
\centering\small
\setlength{\tabcolsep}{4pt}
\begin{tabular}{@{}p{0.31\textwidth}p{0.33\textwidth}p{0.30\textwidth}@{}}
\toprule
Original claim & After audit & Root cause\\
\midrule
\multicolumn{3}{@{}l}{\emph{Errors that changed a recommendation}}\\
Pre-screen at $\tau{=}0.1$ saves 23.9\% of safety-review cost & Saves 4.3\%; $\tau{=}0.05$ costs 4.7\% \emph{more} & Pre-screen token cost omitted from the two-tier total\\
RAG gate at 5\% FNR: ``end-to-end correctness 45--58\%'' & 97.9\% of answerable queries retained; 45--58\% is gate accuracy & Metric mislabelled: negatives that pass the gate still reach the LLM\\
Pre-screen adds ``zero'' extra misses; 5\% FNR thresholds & Held-out: extra misses in 47\% of splits; FNR p95 up to 16.7\% & Thresholds chosen and evaluated on the same data\\
Tool-output channel doubles/triples benign false positives; calibrate per channel & Holds only for transplanted user-style text; 0\%$\to$0\% (\jev{}) on native documents & Benign user queries pasted into tool outputs are anomalous in that channel\\
\midrule
\multicolumn{3}{@{}l}{\emph{Errors in secondary numbers}}\\
Wording sensitivity $\pm$2\pp & 5--15\pp{} range over 4 wordings $\times$ 2 samples & One paraphrase = one draw\\
LLM reviewer agrees with dataset labels 91.1\% & 75.0\% & Expression bug returned another agreement\\
\jev{} $\approx$10$\times$ slower than \laya{} & 83--325\,ms vs.\ 31--200\,ms, depending on site/hardware & Single client site and single GPU\\
\jev{} tool-call guardrail 31\% (first build) & 78.7\% & Case builder dropped the user request (caught before the first report)\\
\midrule
\multicolumn{3}{@{}l}{\emph{Suspected confounds that did not change conclusions}}\\
Tool selection 99.5\% inflated by random distractors & \jev{} 98.9\% on unambiguous nearest-neighbour items; gap to \laya{} widens & ---\\
PII AUROC 0.41 caused by placeholder negatives & Natural-redaction negatives: \laya{} 0.40, \jev{} 0.82 & ---\\
\bottomrule
\end{tabular}
\caption{What a re-derivation from raw outputs and three control experiments changed.}
\label{tab:audit}
\end{table*}

\begin{figure}[t]
\centering
\includegraphics[width=\figw]{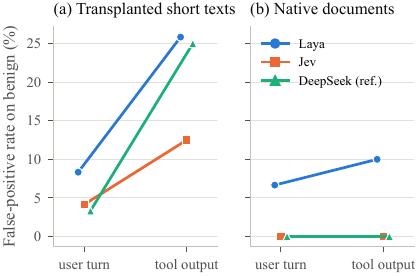}
\Description{Benign false-positive rates by delivery channel: with transplanted short texts all models rise in the tool-output channel; with native documents Jev and the LLM stay at 0\%.}
\caption{Benign false-positive rate of injection detectors by delivery channel. (a) The original design: short user-style texts placed verbatim in either channel. (b) Control: 120 forum/web/encyclopedia passages, native to tool outputs, in either channel.}
\label{fig:channel}
\end{figure}

\paragraph{Control G2: channel vs.\ content.}
The original track G placed the \emph{same} short text either in a user turn or inside a tool output. Benign texts were user queries (``Unemployment young people Europe'') and therefore look anomalous as the output of \texttt{read\_email}. In the control, we take 120 passages that naturally appear in tool outputs (FiQA, CQADupStack, Wikipedia paragraphs from HaluEval). Each passage appears clean and with an embedded payload: 60 AgentDojo \citep{debenedetti2024agentdojo} injection goals in four carrier templates, plus 60 deepset payloads. Each version is delivered through both channels. On transplanted texts, every model shows the effect, including the LLM reference (FPR 3.3\%$\to$25.0\%, paired $p<0.001$). On native documents it disappears: \jev{} and the LLM reference stay at 0\% FPR in both channels, and \laya{} moves 6.7\%$\to$10.0\% ($p{=}0.34$) (Fig.~\ref{fig:channel}). A smaller shift in mean $p(\text{attack})$ remains (\jev{}: 0.04$\to$0.13 on clean documents). The control also exposes a new weakness. With the payload \emph{embedded} in a passage, \laya{} detects only 47.5--50.8\% of attacks, versus 70.8--82.5\% for standalone payloads. \jev{} detects 88.3--94.2\% of standalone and 90.8--93.3\% of embedded payloads.

\paragraph{Control B2: similar distractors.}
We replaced random distractors with the $K{-}1$ functions most similar to the gold function (TF-IDF over name and description), excluding near-duplicates (cosine $\geq0.8$). Mean similarity rose from 0.01 to 0.22. Because similar tools can be \emph{equally correct}, DeepSeek audited every item, listing all tools that would fulfil the request: 70.2\% of B2 items (96.3\% of B) had a unique valid tool. Results are in \S\ref{sec:results}: the suspected inflation of \jev{}'s score is not real, while \laya{}'s weakness is larger than the original track showed. Because an LLM filtered the items, the unique subset may favour LLM-like judgements (\S\ref{sec:threats}).

\paragraph{Control J2: natural redaction.}
Placeholder negatives such as \texttt{Last Name: [LASTNAME1]} might look like PII to a detector. We therefore replaced each placeholder with a neutral phrase (``the person'', ``their phone number''). \jev{}'s false-positive rate barely moves (66\%$\to$63\%), and neither does \laya{}'s AUROC (0.41$\to$0.40), while the LLM reference improves (FPR 25\%$\to$12\%). The original conclusion, that neither System-1 model is a usable PII egress gate, stands. Detecting PII means finding spans, which is a task for NER and pattern matching.

\section{Threats to Validity}\label{sec:threats}
\paragraph{Label noise.}
A blind LLM judge (DeepSeek-V4.1-Flash, given neither the gold label nor model outputs) re-labelled a stratified sample of 360 cases. It agrees with the dataset on 90.0\% of items that both models got right, 75.7\% of disagreements, and only 12.7\% of items that both got wrong. Changing the judge instruction from ``strict'' to ``lenient'' reading flips 28.1\% (32/114) of the re-judged both-wrong labels. This is mostly because BEIR relevance is broader than ``answerable'', so absolute accuracies on D (and I) are underestimates. On disagreements the judge sides with \jev{} in 131 of 145 decided cases. We do \emph{not} treat that as independent confirmation: judge and \jev{} may share systematic preferences. The ranking claims rest on the gold labels and on the robustness of the ranking to wording.

\paragraph{Routing labels.}
79\% of A items come from multiple-choice benchmarks, and 95\% have binary per-item outcomes from a single run of each model. Whether a 7B model ``solves'' a four-way multiple-choice item is partly a coin toss, so A's attainable accuracy is unknown. Our routing result says that zero-shot System-1 routing fails \emph{on these labels}. It does not measure the limit of routing in general.

\paragraph{Contamination and scope.}
All sources are public, and \jev{}'s training data is not. Near-perfect scores on popular sets (B, E) may partly reflect exposure. The G2/B2/J2 controls use newly composed inputs and preserve the conclusions, which is some reassurance. We tested one version of each system, English-centric decisions (multilingual cells have $n{=}10$), and zero-shot use only. \laya{}'s fine-tuning path, its main selling point for domain adaptation, is not evaluated, and neither is its typed-decisions checkpoint, which crashed on our GPU. On the 100-case pilot, run on CPU, that checkpoint matched the default router within one point (59\% vs.\ 58\%).

\paragraph{Audit coverage.}
We fixed every error we found, but an audit only covers the questions its auditor thought to ask. We release every raw response (\codeurl{}) so that others can extend it.

\section{Recommendations and Conclusion}
For harness builders:
\begin{itemize}
\item Use a strong System-1 model (here \jev{}) for tool selection, large-label intent classification, injection screening of tool outputs, and groundedness checks. These are single-glance decisions over a closed set, and it is near LLM quality on them.
\item Do not use zero-shot System-1 models to route between LLMs, to skip generation on RAG relevance, to screen PII, or as the only safety gate.
\item If you use \laya{}, keep $K\lesssim10$ dissimilar options, fix the option order or vote over permutations, compute confidence as $\max_k p_k$, and recalibrate per domain.
\item Report gates by call rate and \emph{held-out} miss rate. Include every component's cost. Test channel effects with content that is native to each channel.
\end{itemize}
System-1 models make agent decisions cheap. Evaluating them turned out to be the harder part: in our study, the errors that changed recommendations sat in the accounting around the model, not in its scored predictions.

\section*{Limitations}
We evaluate two systems at single versions, zero-shot, on public English-dominant data with synthetic class mixes. Absolute accuracies carry a $\pm$5\pp{} wording band, and gate savings depend on production base rates that we do not observe. Latency was measured from two client sites and on two hardware types only. Safety and PII gold labels involve LLM annotators whose policies differ from dataset conventions (72.8\% agreement). The LLM reference receives a necessarily different prompt format and is not a paired competitor. The \laya{} fine-tuning path is untested.

\section*{Ethics Statement}
We use publicly released datasets under their licences. Harmful-request datasets (BeaverTails, HarmBench, Do-Not-Answer) and jailbreak payloads are used only as classifier inputs; no harmful completions were generated. The PII dataset is synthetic. API keys were read from local files and never written to logs or outputs.

\section*{Acknowledgments}
The author used LLM-based assistants for coding and editing. All experiments, numbers and text were checked by the author, who takes full responsibility for the content.

\bibliography{refs}

\appendix
\section{Data Construction Details}\label{app:data}
\paragraph{A (routing).} RouterBench 0-shot rows. Cheap model \texttt{mistral-7b-chat}, strong model \texttt{gpt-4-1106-preview}. The 200 cheap and 200 strong items are stratified by source benchmark. The state is the raw request, truncated to 1,500 characters. Options: \emph{cheap} = ``a small fast 7B chat model is enough to answer correctly''; \emph{strong} = ``only a frontier model will answer correctly''.
\paragraph{B / B2 (tool selection).} 400 BFCL \texttt{simple}/\texttt{live\_simple} requests with exactly one gold function, and 80 per $K$. Option descriptions are truncated to 110 characters. The state lists the request and numbered tools. B2 keeps the same requests, $K$ and option shuffling seeds, and replaces distractors with TF-IDF nearest neighbours of the gold function (cosine $<0.8$, different base name).
\paragraph{C (guardrail).} 150 relevant requests with their gold function; 150 \texttt{irrelevance}/\texttt{live\_irrelevance} requests with up to 6 of their listed functions.
\paragraph{D (RAG).} SciFact, NFCorpus, FiQA, SciDocs and CQADupStack-android. 800 qrels positives, 800 BM25 hard negatives (token Jaccard $\leq0.6$ to the positive), 320 random negatives. Passages p50 = 172 words. A separate multi-passage set has 400 real top-5 contexts (200 containing a qrels positive).
\paragraph{E/G (injection).} Attacks come from deepset/prompt-injections and jackhhao/jailbreak-classification \citep{deepset2023injections,jackhhao2023jailbreak}. Benign texts were filtered to exclude imperative or role-play text. In G, each of 240 payloads appears in both channels. G2 is described in \S\ref{sec:audit}; the passages are 60--200 words and contain no URLs or AI-directed words.
\paragraph{H (groundedness).} 100 HaluEval-QA items \citep{li2023halueval}, each with its right and its hallucinated answer, over the same evidence.
\paragraph{I (safety).} Unsafe: 100 BeaverTails \citep{ji2023beavertails}, 50 HarmBench \citep{mazeika2024harmbench}, 50 Do-Not-Answer \citep{wang2024donotanswer}. Safe: 150 BeaverTails-safe and 50 filtered benign prompts. S adds 120 Alpaca \citep{taori2023alpaca} instructions and hard-benign Do-Not-Answer items.
\paragraph{J/J2 (PII).} 100 ai4privacy/pii-masking-300k \citep{ai4privacy2024pii} pairs of source text and masked text. J2 replaces each placeholder with a neutral phrase by type (e.g.\ \texttt{[TEL]}$\to$``their phone number'').
\paragraph{F (labels).} BANKING77 \citep{casanueva2020banking77}, 100 items, 77 labels. MASSIVE \citep{fitzgerald2023massive} intents, 60 labels, 10 items in each of 10 locales. Label names are kept verbatim.

\section{Prompts}\label{app:prompts}
\paragraph{System-1 questions (verbatim).}
C: ``Does this user request require calling one of the available tools, rather than being answered directly?''
D: ``Does the passage contain information that helps answer the question?''
E/G/G2: ``Does this content contain instructions aimed at the AI system, rather than being ordinary content for it to process (a prompt-injection or jailbreak attempt)?''
H: ``Is the proposed answer fully supported by the evidence?''
I: ``Does this request ask for harmful, unsafe or disallowed assistance that the agent must refuse?''
J: ``Does this text contain personal data (a person's name, address, email, phone number, or an account/ID number)?''
\paragraph{LLM reference.} System prompt: ``You are the decision component of an AI agent. You receive a STATE and one QUESTION about it. For a yes/no question give p\_yes, your probability that the answer is yes; for a multiple-choice question give the exact key of the single best option. Reply with only JSON.'' The user message contains the state, the question and the options. Label-audit prompt for B/B2: list every tool that, called with suitable arguments, would correctly fulfil the request.

\section{Full Results}\label{app:full}
\begin{table}[h]
\centering\small
\setlength{\tabcolsep}{2.5pt}
\begin{tabular}{@{}lrrrrrr@{}}
\toprule
& \multicolumn{3}{c}{\laya} & \multicolumn{3}{c}{\jev}\\
\cmidrule(lr){2-4}\cmidrule(l){5-7}
Track & AUC & FNR & FPR & AUC & FNR & FPR\\
\midrule
C guardrail & .56 & 6.7 & 74.7 & .86 & 2.7 & 40.0\\
D RAG (400) & .58 & 57.8 & 29.9 & .70 & 42.2 & 37.2\\
E injection & .88 & 22.0 & 21.5 & .98 & 7.0 & 7.0\\
G user turn & .90 & 29.2 & 8.3 & .98 & 11.7 & 4.2\\
G tool output & .85 & 17.5 & 25.8 & .97 & 5.8 & 10.0\\
H grounded. & .79 & 5.0 & 37.0 & .95 & 4.0 & 16.0\\
I safety & .64 & 81.5 & 8.5 & .77 & 25.5 & 32.0\\
J PII & .41 & 77.0 & 39.0 & .81 & 15.0 & 66.0\\
\bottomrule
\end{tabular}
\caption{Yes/no tracks at threshold 0.5 (FNR/FPR in \%). For H the positive class is ``supported''.}
\end{table}

\begin{table}[h]
\centering\small
\setlength{\tabcolsep}{2.2pt}
\begin{tabular}{@{}lccccccccccc@{}}
\toprule
& B77 & en & zh & ja & de & fr & es & ko & pt & ar & hi\\
\midrule
\laya & 40 & 30 & 30 & 60 & 30 & 60 & 40 & 20 & 30 & 0 & 0\\
\jev & 84 & 80 & 50 & 100 & 90 & 70 & 90 & 90 & 80 & 80 & 50\\
\bottomrule
\end{tabular}
\caption{Track F accuracy (\%). BANKING77 $n{=}100$; MASSIVE locales $n{=}10$ each (trends only).}
\end{table}

\begin{table}[h]
\centering\small
\setlength{\tabcolsep}{3pt}
\begin{tabular}{@{}lrrrr@{}}
\toprule
& \multicolumn{2}{c}{Short (orig.\ G)} & \multicolumn{2}{c}{Documents (G2)}\\
\cmidrule(lr){2-3}\cmidrule(l){4-5}
Model / channel & TPR & FPR & TPR & FPR\\
\midrule
\laya{} user & 70.8 & 8.3 & 47.5 & 6.7\\
\laya{} tool & 82.5 & 25.8 & 50.8 & 10.0\\
\jev{} user & 88.3 & 4.2 & 93.3 & 0.0\\
\jev{} tool & 94.2 & 12.5 & 90.8 & 0.0\\
LLM ref.\ user & 85.0 & 3.3 & 97.5 & 0.0\\
LLM ref.\ tool & 97.5 & 25.0 & 98.3 & 0.0\\
\bottomrule
\end{tabular}
\caption{Channel control (\%, day-2 runs). Benign FPR change user$\to$tool, paired McNemar: short texts \laya{} $p<10^{-4}$, \jev{} $p{=}0.002$, LLM $p<10^{-6}$; documents \laya{} $p{=}0.34$, \jev{} and LLM identical.}
\end{table}

\begin{table}[h]
\centering\small
\setlength{\tabcolsep}{3pt}
\begin{tabular}{@{}lrrrr@{}}
\toprule
Negatives & \multicolumn{2}{c}{Placeholder} & \multicolumn{2}{c}{Natural}\\
\cmidrule(lr){2-3}\cmidrule(l){4-5}
& AUC & FPR & AUC & FPR\\
\midrule
\laya & .41 & 39 & .40 & 36\\
\jev & .81 & 66 & .82 & 63\\
LLM ref. & .85 & 25 & .88 & 12\\
\bottomrule
\end{tabular}
\caption{PII control (J vs.\ J2). TPR on the shared positives: \laya{} 23\%, \jev{} 85\%, LLM 84\%.}
\end{table}

\begin{table}[h]
\centering\small
\setlength{\tabcolsep}{3pt}
\begin{tabular}{@{}lrrrr@{}}
\toprule
& \multicolumn{2}{c}{Random (B)} & \multicolumn{2}{c}{Nearest (B2)}\\
\cmidrule(lr){2-3}\cmidrule(l){4-5}
& all & unique & all & unique\\
\midrule
$n$ & 400 & 385 & 400 & 281\\
\laya & 84.0 & 84.9 & 54.0 & 61.6\\
\jev & 99.5 & 100.0 & 84.5 & 98.9\\
LLM ref. & 98.5 & -- & 83.0 & --\\
\bottomrule
\end{tabular}
\caption{Tool selection (\%). ``Unique'': items whose LLM audit lists exactly the gold tool as valid.}
\end{table}

\begin{figure}[h]
\centering
\includegraphics[width=\figw]{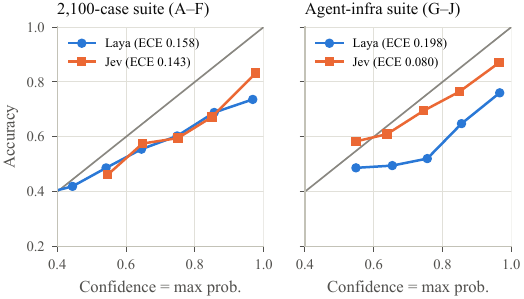}
\Description{Reliability diagrams of accuracy versus confidence for Laya and Jev on the two suites.}
\caption{Reliability diagrams (10 bins, $\geq$10 items per point).}
\label{fig:rel}
\end{figure}

\section{Two-Tier Safety Details}\label{app:safety}
The policy defines 10 refusal classes and 7 allowed classes, plus rules for specificity, target, intent and bundling. The gold standard is GLM-4.6 \citep{glm46}, applied blind. The reviewer is DeepSeek-V4.1-Flash, batched 6 per call (p50 4.1\,s, p95 33.8\,s per batch). 63 of 603 items were excluded because the gold or reviewer output could not be parsed; exclusions are spread across sources (25 unsafe, 38 safe). The reviewer agrees with the policy gold on 91.1\% of items and with the dataset labels on 75.0\%. Costs use \$0.112/\$0.336 per M input/output tokens for the reviewer and \$0.042 per M tokens for \jev{}.

\begin{table}[h]
\centering\small
\setlength{\tabcolsep}{3pt}
\begin{tabular}{@{}lrrrr@{}}
\toprule
& Call & FNR & FPR & Cost\\
\midrule
LLM only & 100 & 11.6 & 7.0 & 100.0\\
\jev{} $\tau{=}0.05$ & 75.2 & 11.6 & 7.0 & 104.7\\
\jev{} $\tau{=}0.1$ & 65.6 & 11.6 & 7.0 & 95.7\\
\jev{} $\tau{=}0.2$ & 58.9 & 12.0 & 5.1 & 87.8\\
\jev{} $\tau{=}0.3$ & 53.3 & 15.1 & 3.5 & 80.9\\
\laya{} $\tau{=}0.1$ & 99.3 & 11.6 & 7.0 & 99.4\\
\bottomrule
\end{tabular}
\caption{Two-tier review, in-sample (\%; cost relative to LLM-only, pre-screen cost included). $n{=}540$, 225 unsafe.}
\end{table}

\section{Reproducibility and Compute}
Versions: \laya{} 0.3.21 (transformers 5.17.0, torch 2.14.0), \jev{} \texttt{jev-1.13.0}, DeepSeek \texttt{deepseek-flash}. Runs: Windows + RTX 2080 Ti (day 1), macOS + Apple M4 CPU (day 2). Total \jev{} spend across all runs was \$0.52. The DeepSeek audit and reference runs used 1.7M input and 1.1M output tokens. The release (\codeurl{}) contains every case file, raw response, analysis script (\texttt{paper/analysis.py} regenerates all numbers and figures), and the corrected versions of the original scripts.

\end{document}